\documentclass[10pt,twocolumn,letterpaper]{article}

\usepackage{wacv}
\usepackage{times}
\usepackage{epsfig}
\usepackage{graphicx}
\usepackage{amsmath}
\usepackage{amssymb}
\usepackage{booktabs}
\usepackage{appendix}
\usepackage{arydshln}

\usepackage{import}
\usepackage{xcolor}         
\usepackage{color,soul}     
\usepackage{makecell}
\usepackage{subcaption}
\usepackage{array}%
\usepackage{multirow}
\usepackage{enumitem}
\usepackage{graphicx}
\usepackage[accsupp]{axessibility}  

\def\wacvPaperID{1172} 

\wacvalgorithmstrack   

\wacvfinalcopy 

\ifwacvfinal
\usepackage[breaklinks=true,bookmarks=false]{hyperref}
\else
\usepackage[pagebackref=true,breaklinks=true,colorlinks,bookmarks=false]{hyperref}
\fi

\hypersetup{
    colorlinks=true,
    linkcolor=blue,
    filecolor=magenta,      
    urlcolor=blue,
    citecolor=blue,
}

\definecolor{blue_munsell}{rgb}{0.36, 0.54, 0.66}
\definecolor{blue-violet}{rgb}{0.54, 0.17, 0.89}
\definecolor{byzantine}{rgb}{0.74, 0.2, 0.64}
\definecolor{caputmortuum}{rgb}{0.35, 0.15, 0.13}

\newcommand*{\imagenet}{\textsc{ImageNet}\xspace}
\newcommand*{\imagenetk}{\textsc{ImageNet-21k}\xspace}
\newcommand*{\cifar}{\textsc{CIFAR100}\xspace}
\newcommand*{\places}{\textsc{Places365}\xspace}
\newcommand*{\csaw}{\textsc{CSAW}\xspace}

\newcommand*{\deit}{\textsc{DeiT}\xspace}
\newcommand*{\deits}{\textsc{DeiT}s\xspace}
\newcommand*{\deitsmall}{\textsc{DeiT-S}\xspace}
\newcommand*{\deitTiny}{\textsc{DeiT-T}\xspace}
\newcommand*{\deitBase}{\textsc{DeiT-B}\xspace}
\newcommand*{\deitLarge}{\textsc{DeiT-L}\xspace}

\newcommand*{\swins}{\textsc{SWIN}s\xspace}
\newcommand*{\cls}{{\fontfamily{qcr}\selectfont[CLS]}\xspace}

\begin{document}

\title{PatchDropout: Economizing Vision Transformers Using Patch Dropout}

\author{\noindent
{Yue Liu} $^{1,2}$\textsuperscript{
\thanks{Corresponding author: Yue Liu \textless{}yue3@kth.se\textgreater{}}}, 
\vspace{1mm}
{Christos Matsoukas} $^{1,2,5}$, 
{Fredrik Strand} $^{3,4}$,
{Hossein Azizpour} $^{1}$, 
{Kevin Smith} $^{1,2}$\\[2mm]
\small \vspace{-1mm}
{\noindent
$^{1}$ KTH Royal Institute of Technology, Stockholm, Sweden} 
\small
$^{2}$ Science for Life Laboratory, Stockholm, Sweden \\
\small \vspace{-1mm}
$^{3}$ Karolinska Institutet, Stockholm, Sweden 
\small 
$^{4}$ Karolinska University Hospital, Stockholm, Sweden \\
\small \vspace{-1mm}
$^{5}$ AstraZeneca, Gothenburg, Sweden \\
}


\maketitle
\thispagestyle{empty}


\begin{abstract}

Vision transformers have demonstrated the potential to outperform CNNs in a variety of vision tasks. 
But the computational and memory requirements of these models prohibit their use in many applications, especially those that depend on high-resolution images, such as medical image classification.
Efforts to train ViTs more efficiently are overly complicated, necessitating architectural changes or intricate training 
schemes.
In this work, we show that standard ViT models can be efficiently trained at high resolution by randomly dropping input image patches.
This simple approach, PatchDropout, reduces FLOPs and memory by at least 50\% in standard natural image datasets such as \imagenet, and those savings only increase with image size.
On \csaw, a high-resolution medical dataset, we observe a $5\times$ savings in computation and memory using PatchDropout, along with a boost in performance.
For practitioners with a fixed computational or memory budget, PatchDropout makes it possible to choose image resolution, hyperparameters, or model size to get the most performance out of their model.

\end{abstract}
\section{Introduction}


\begin{figure}[t!]
\begin{center}
\vspace{-1mm}
\includegraphics[width=1\columnwidth]{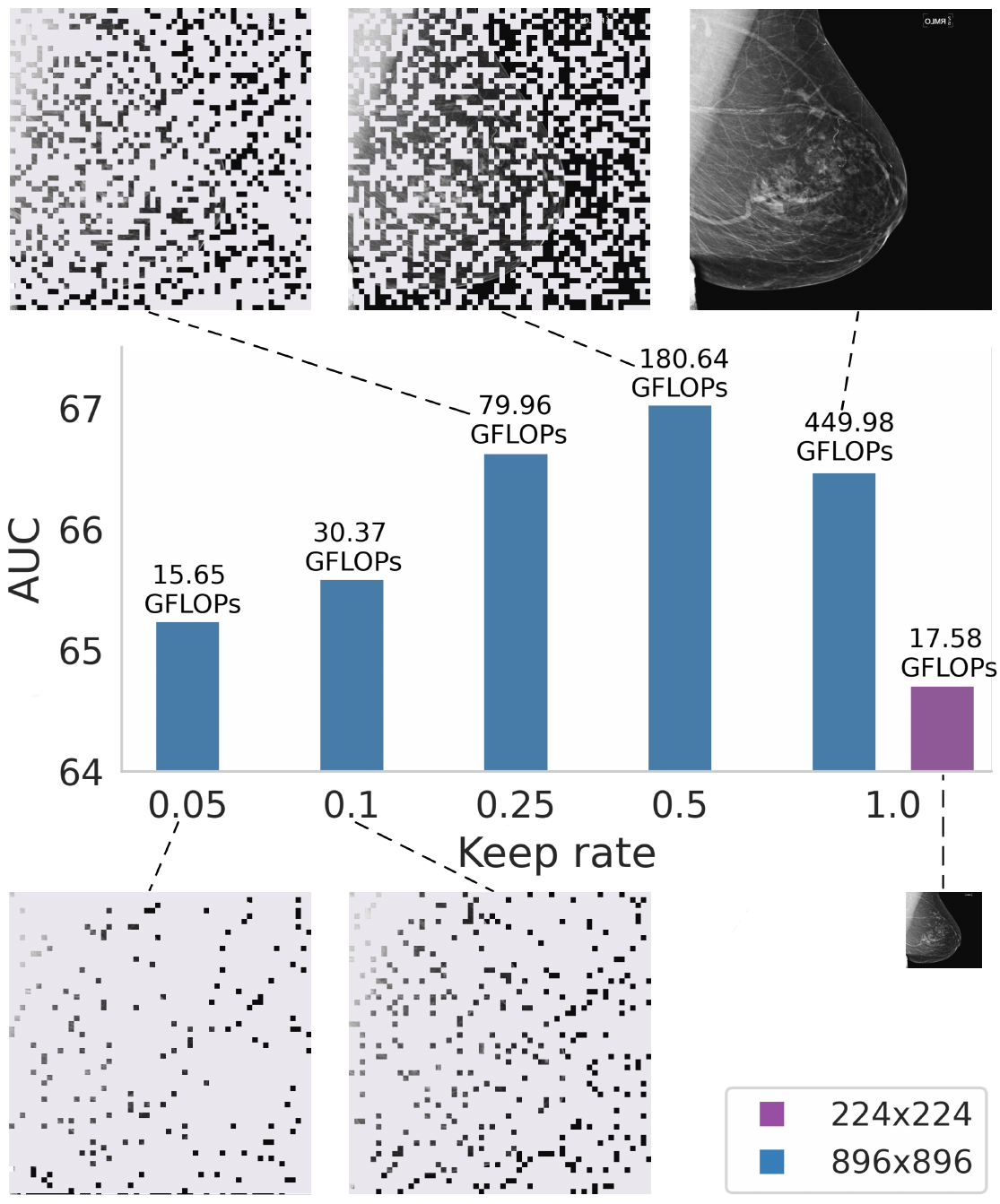}
\end{center}
\vspace{-2mm}
\caption{
    \emph{PatchDropout can be used to efficiently train off-the-shelf ViTs on high-resolution images. 
    }
    A tiny proportion of the input patches is often enough for accurate prediction, and if larger resolution images are used performance can increase, despite the lost information. 
    Here, training with different ratios of input patches affects the model's performance on \csaw, a real-world medical dataset with high resolution images.
    PatchDropout results in increased predictive performance under the same computational budget.
    Using a $16 \times$ larger image but keeping $5\%$ of the patches saves computation and memory, and improves performance.
    Further improvements can be achieved by increasing the keep rate, at the expense of computation.
    A similar trend is observed across standard image classification datasets.
    }
\label{fig:fig1}
\vspace{-2mm}
\end{figure}

Vision Transformers (ViTs) \cite{dosovitskiy2020image} have been recently introduced as a viable alternative to CNNs \cite{dosovitskiy2020image, liu2021swin, matsoukas2021time, shamshad2022transformers}.
However, promises of better performance have not yet been realized in many settings due to computational bottlenecks.
For instance, ViTs require large datasets to train on \cite{dosovitskiy2020image}, though this issue has been partially solved using pre-training on large datasets  \cite{dosovitskiy2020image,caron2021emerging}.
Memory and compute requirements add to this, since the self-attention mechanism introduces an element with quadratic complexity \emph{w.r.t.}~the number of tokens.
These bottlenecks can result in long training times, and for large images such as those encountered in medical image analysis, the computational and memory demands render off-the-shelf ViTs unsuitable.


These computational issues are acute in other domains as well, \textit{e.g.}~microscopy and remote sensing, especially when native resolution is not only a desired property but a requirement for accurate predictions.
Accordingly, several works focus on making vision transformers more efficient using a plethora of different approaches, which usually involve some kind of post-processing or architectural modifications \cite{rao2021dynamicvit, wang2021not, yang2021nvit, tang2022patch, yin2022vit}.
These methods prioritize efficiency during inference, \emph{e.g.}~for embedding in mobile devices, and have been shown to reduce run-time by 30 to 50\% without compromising performance.
However, the bottleneck in network training can not be overlooked. 
Few works have addressed this topic, and those that do require architecture modifications or complex training schemes which limits their use \cite{wang2020linformer,liu2021swin,wang2021pyramid,renggli2022learning,ryoo2021tokenlearner}.
Efficient ViT training remains an important problem, especially for applications requiring large images, as all but the largest institutions are limited by computational resources to train ViTs.



In this work, we ask a fundamental question.
\textit{Are all input patches necessary during training, or can we randomly disregard a large proportion of them?}
An affirmative response entails a simple, yet efficient approach that reduces compute and memory footprint.
Our method, PatchDropout, randomly drops input tokens and results in
up to $5\times$ reduction in memory and compute during training when using high-resolution images, without compromising the model accuracy (Figure \ref{fig:fig1}).
This can be achieved with off-the-shelf vision transformers and a minimal implementation, owing to the nature of ViTs.
Furthermore, we show that given a fixed memory and computational budget, PatchDropout makes it possible to choose image resolution, hyperparameters, or model
size to get the most performance out of the model.
We conduct experiments on \csaw, a real-world medical dataset with high resolution images, and further validate our proposed method using three mainstream datasets: \imagenet, \cifar and \places.
Through these experiments we show that:

\begin{itemize}
    \vspace{-1mm}
    \item We can randomly discard image patches during training without compromising performance and improve efficiency from $2 \times$ up to $5.6 \times$, depending on image size (see Figures \ref{fig:fig1} and \ref{fig:main_results}).
    \vspace{-1mm}
    \item Given the same computational budget, up-scaling the images and/or utilising a larger ViT variant while discarding a fraction of input tokens can improve the model's accuracy (see Table \ref{tab:token_number} and Table \ref{tab:model_variant}).
    \vspace{-1mm}
    \item PatchDropout can act as a regularization technique during training, resulting in increased model robustness (see Figure \ref{fig:robustness}).
\end{itemize}

These findings along with additional ablation studies suggest that PatchDropout can economize ViTs, allowing their utilization on high-resolution images, with potential gains in accuracy and robustness.
Code to reproduce our work is available at \href{https://github.com/yueliukth/PatchDropout}{https://github.com/yueliukth/PatchDropout}.

\section{Related Work}
\label{related}


Several studies have examined how to obtain a lighter vision transformer model using an existing well-trained one to improve inference efficiency using \emph{e.g.}~pruning or a teacher for distillation.
DynamicViT \cite{rao2021dynamicvit} adds a prediction module for estimating the importance score of each patch progressively. The training is assisted by knowledge distillation and the patches whose contribution are minimal to the final prediction are pruned during inference. 
Another pruning method, PatchSlimming \cite{tang2022patch} identifies less important patches from the last layer and removes them from previous ones. 
DVT \cite{wang2021not} dynamically determines the number of patches by training a cascade of transformers using an increasing number of patches and then interrupts inference once the prediction is confident. 


Another line of research focuses on making the training more efficient. 
Several studies attempt optimization of network architectures through artificially designed modules \cite{wang2020linformer,liu2021swin,wang2021pyramid}, among which PatchMerger \cite{renggli2022learning} and TokenLearner \cite{ryoo2021tokenlearner} are designed specifically for reducing the number of tokens. 
EViT \cite{liang2022not} learns to gradually preserve the attentive tokens and fuse the inattentive ones during training which results in a 0.3\% decrease in accuracy on \imagenet with a 50\% increase in speed of inference. 
Compared to EViT, the proposed method of this study is complementary but with a much simpler mechanism that does not require substantial modifications. 


A few recent works explore the possibility of learning expressive representations by selecting a subset of patches. 
MAE \cite{he2022masked}, which is designed for more efficient self-supervised pre-training, proposes dropping a high proportion of patches and subsequently inferring the missing patches through an autoencoder setup. Our work takes some inspiration from MAE, however, PatchDropout can be applied to target tasks directly using standard ViTs (unlike MAE).
In \cite{hao2022learnable}, the authors augment standard ViTs with additional patches that selectively attend to a subset of patches to improve transferability of ViTs.
Finally, \cite{naseer2021intriguing} shows that ViTs are robust to random occlusions. 
However, it should be noted that occlusion does not result in efficiency improvement.


\section{Methods}

\label{methods}



\begin{figure}[t]
\begin{center}
\vspace{-2mm}
\begin{tabular}{@{}l@{\hspace{10mm}}l@{}l@{}}
    \includegraphics[width=0.4\columnwidth]{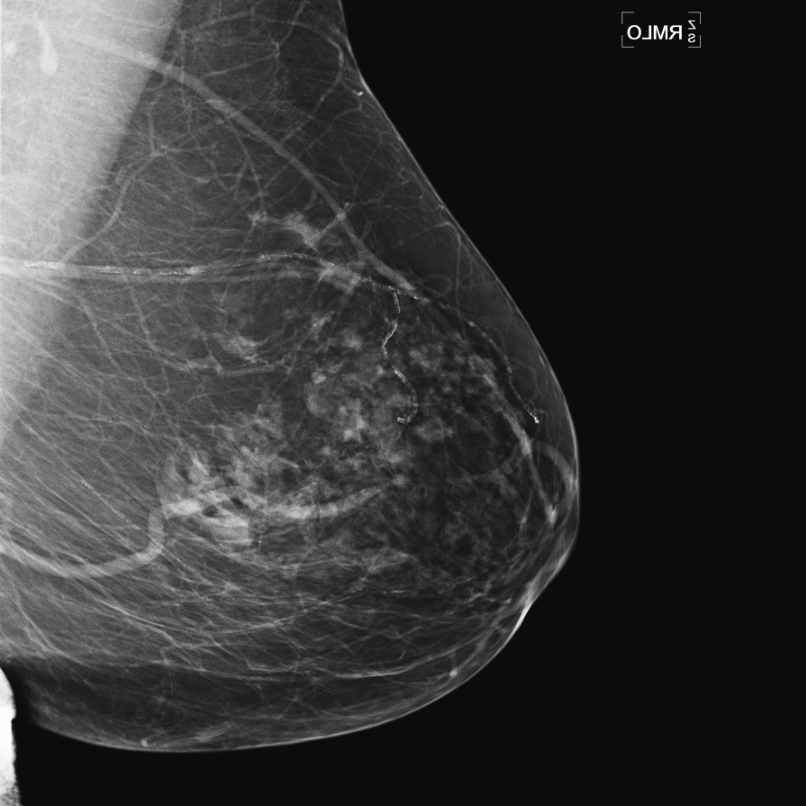} & &
    \includegraphics[width=0.4\columnwidth]{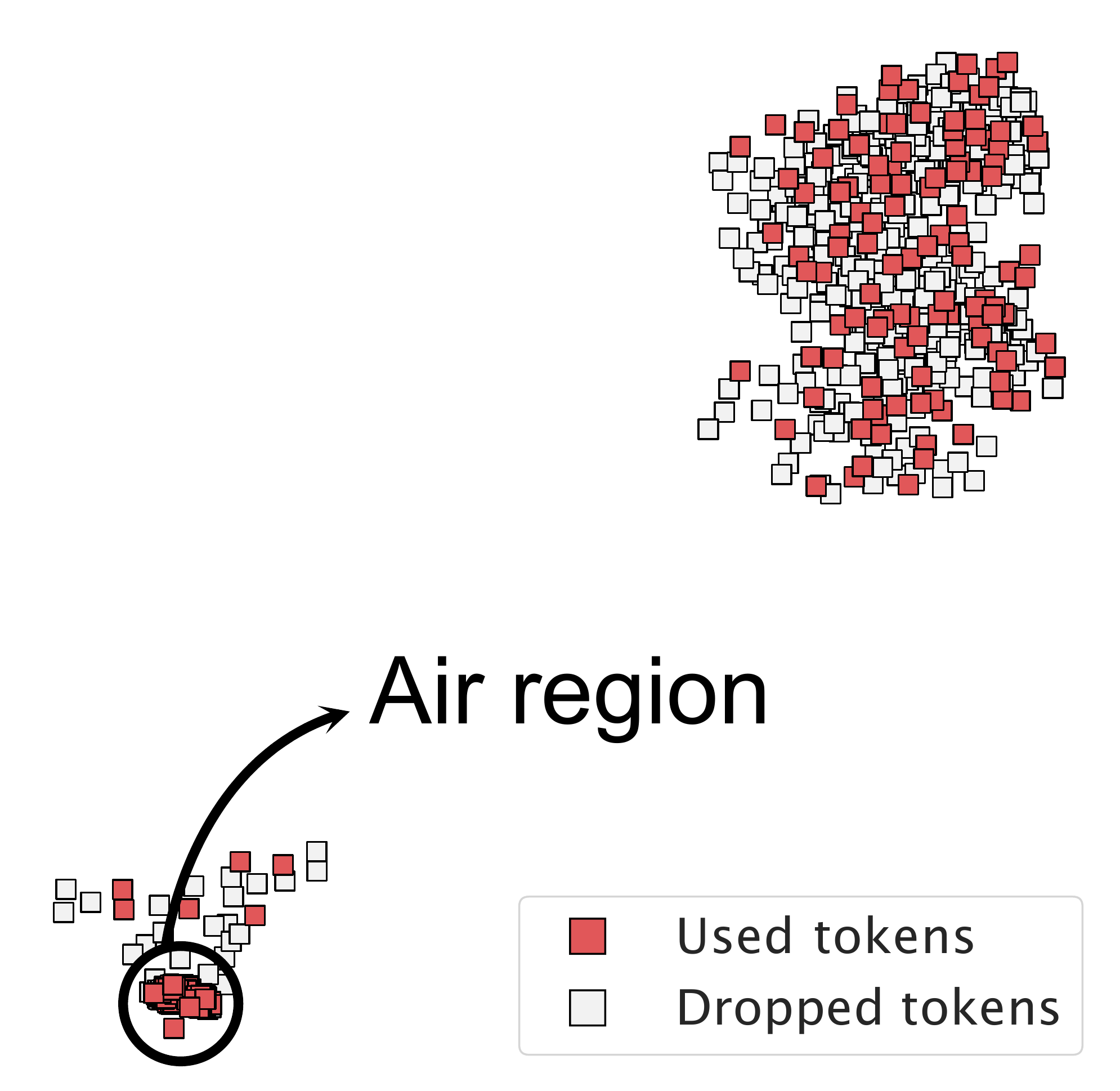}\\
\end{tabular}
\end{center}
\vspace{-3mm}
\caption{
    \emph{Redundancy in mammographic images.}
    \textbf{(Left)} An example image from \csaw.
    \textbf{(Right)} 2D projection of the extracted patches of the left image using UMAP \cite{mcinnes2018umap}. 
    The red squares represent the randomly kept patches after PatchDropout with a keep rate of 0.25.
    The patches are clustered into 2 distinct groups: the air and the breast regions.
    Surprisingly, by simply sampling uniformly, the information needed for accurate classification is retained.
    }
\label{fig:vis_red}
\vspace{-4mm}
\end{figure}

Transformer models were originally developed for language-related tasks \cite{vaswani2017attention}, but their self-attention mechanism has been proven useful for vision tasks as well \cite{dosovitskiy2020image, touvron2021training, liu2021swin}.
An important difference between the two tasks is that visual data, often, contains considerable redundancy or correlation in appearance \cite{he2022masked} (see Figure \ref{fig:vis_red}).
This observation leads to the following question: \textit{Can we randomly omit input image patches during training? If yes, what are the benefits of doing so?}
Here, we aim to answer these questions, showing that vision transformers can indeed be trained using a fraction of the input data and perform well, while at the same time saving a significant amount of memory and compute.
Additionally, our simple training scheme may offer some desirable regularization effects.


\subsection{PatchDropout} \label{method} 

\begin{figure}[t]
\begin{center}
\vspace{-4mm}
\begin{tabular}{@{}c@{}l@{\hspace{2mm}}c@{}}

    Training && Inference \vspace{3mm}
    \\
    \multicolumn{1}{l|}{\includegraphics[width=0.43\columnwidth]{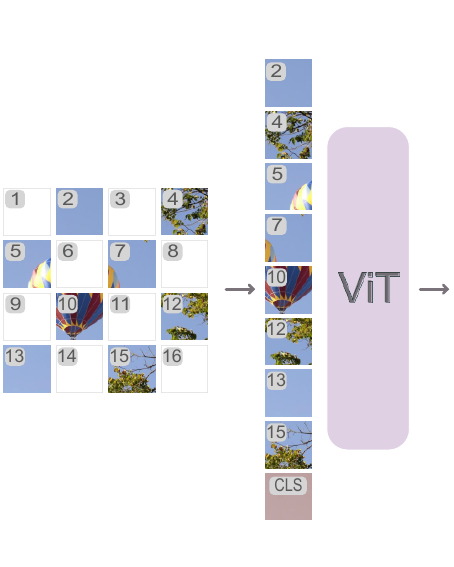}}  & & 
    \includegraphics[width=0.43\columnwidth]{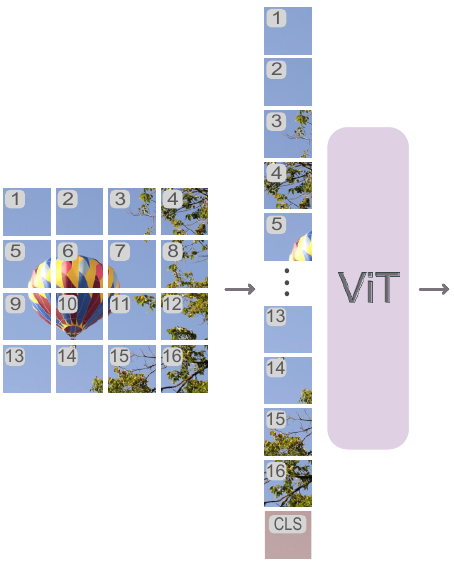}\\
\end{tabular}
\end{center}
\vspace{-4mm}
\caption{
    \emph{PatchDropout during training and inference.}
     \textbf{(Left)} 
    PatchDropout is easy to implement. Patchify the image and add positional embeddings to each patch. Uniformly sample a subset of them and use them to train the model. 
     \textbf{(Right)} At test time, all patches are retained.
    }
\label{fig:vis_patchdropout}
\vspace{-4mm}
\end{figure}

Our core idea relies on the fact that the spatial redundancy encountered in image data can be leveraged to economize vision transformers.
If we randomly deny a fraction of the information to the model during training, we expect a diminished impact on the model's predictive performance.
PatchDropout implements this by randomly dropping a percentage of image tokens at the input level (see Figure \ref{fig:vis_patchdropout}).
More specifically, before the patch embeddings are sent to transformer blocks, a subset of tokens is randomly sampled without replacement. 
Positional embeddings are added prior to the random sampling so that the corresponding position information is retained. The \cls token is retained if it exists. 
The sampled token sequence is sent to transformer blocks in the standard manner.
The proposed method is straightforward and trivial to implement, which makes it viable to be incorporated in most ViT models without substantial modifications.

\subsection{Complexity Analysis}\label{section:complexity}


Vision transformers operate on a series of tokens, where each token corresponds to a non-overlapping image patch and is represented by a linear projection of the patch summed with a positional embedding.
In practice, an image of size $H\times W$ is tiled into $N=HW/P^2$ patches, where $P$ is the patch size and it is typically defined by the user (often $8$ or $16$).
The resulting token sequence is fed into a series of consecutive transformer blocks that update the $d$-dimensional embedding of tokens and consist of a Multi-head Self-Attention (MSA) and Multi-Layer Perceptron modules (MLP).
The MSA itself includes a series of MLP layers that model the interactions between the tokens through attention. 
A final MLP layer is responsible for projecting the output to have the same dimensions as its input, ready to be processed by the next transformer block.
Given this information, we can discuss the theoretical and empirical computational complexity of vision transformers.


\paragraph{Theoretical complexity}
Given $L$ transformer blocks with $N$ tokens and $d$-dimensional embeddings, the computational cost of the self-attention within the MSA module is $\mathcal{O}(LN^2d)$, while the other MLP layers introduce a complexity of $\mathcal{O}(LNd^2)$.
In total, the computational complexity of a series of $L$ transformer blocks is: 
\begin{equation}\label{eq1}
2LN^2d+4LNd^2.
\end{equation} 

The compute is always linear to the depth $L$.
When $N\gg d$ the complexity reduces to the first term, when $N \ll d$ it reduces to the second term. 

For high resolution images with small patch sizes, which is the focus of this work, the first term prevails. 
This leads to a quadratic complexity with respect to the sequence length $N$.
Accordingly, removing a non-trivial portion of the input tokens can result in significant savings in compute.

\paragraph{Empirical complexity}
In practice, the observed computation cost may not exactly reflect the theoretical prediction.
A few factors can make the computational saving of PatchDropout less advantageous than the complexity analysis might suggest. 
For instance, the patchifier, \ie the layer responsible for tokenizing and projecting the input image into a series of embedded tokens adds computational overhead.
The same is true for the classification head.
Nonetheless, as image size increases (and thus the input sequence length $N$ increases), the gap between the theoretical and empirical relative computation should lessen. 
This necessitates an empirical analysis of the savings in computation to confirm the theoretical predictions.

\begin{figure}[t]
\centering
\scriptsize
\begin{tabular}{@{}c@{}c@{}c@{}}
\textbf{Keep rate 0.5 (Theoretical)} &
\textbf{Keep rate 0.5 (Empirical)} &\vspace{2mm}
\\
\includegraphics[width=0.5\linewidth]{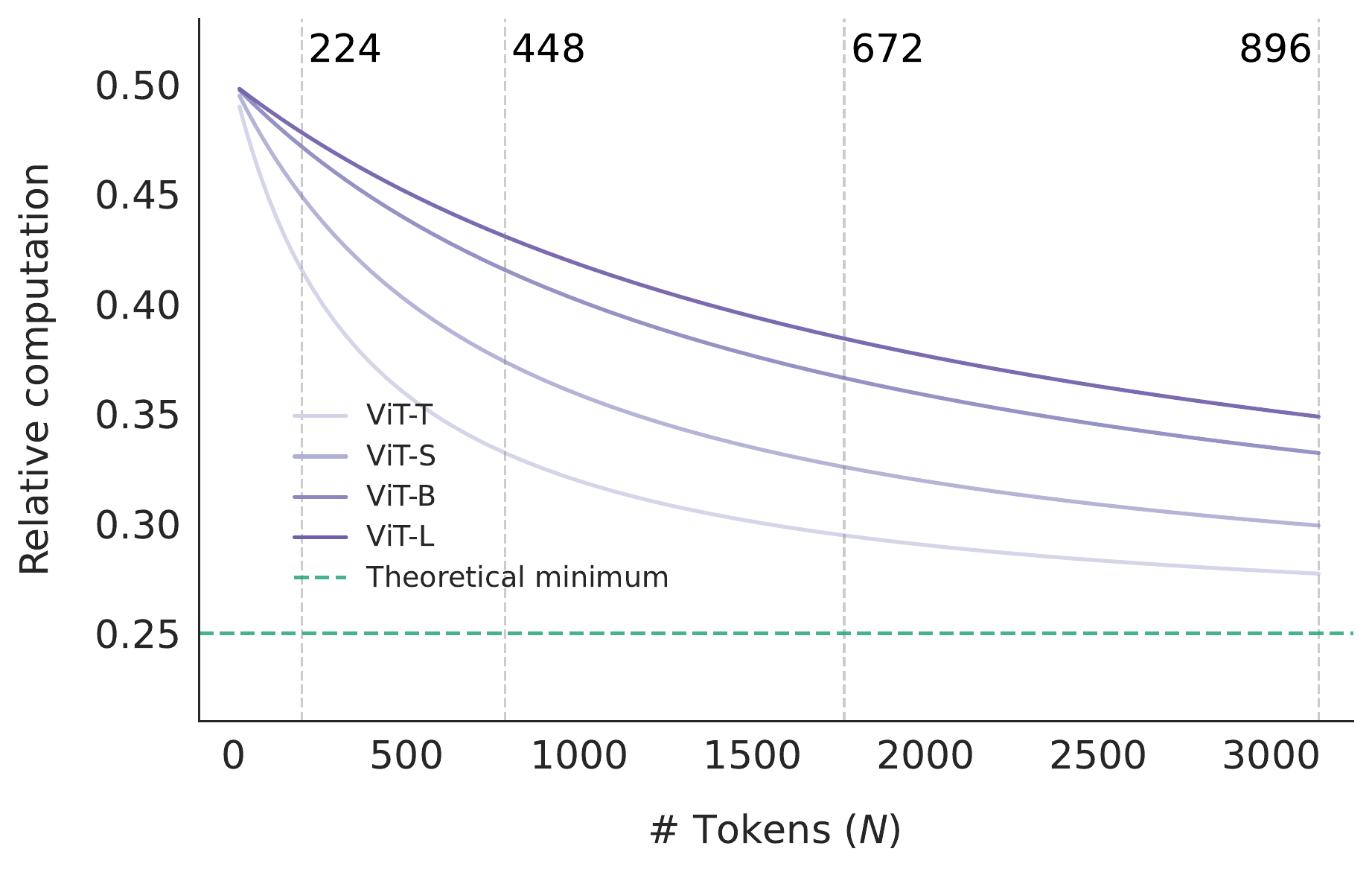} &
\includegraphics[width=0.5\linewidth]{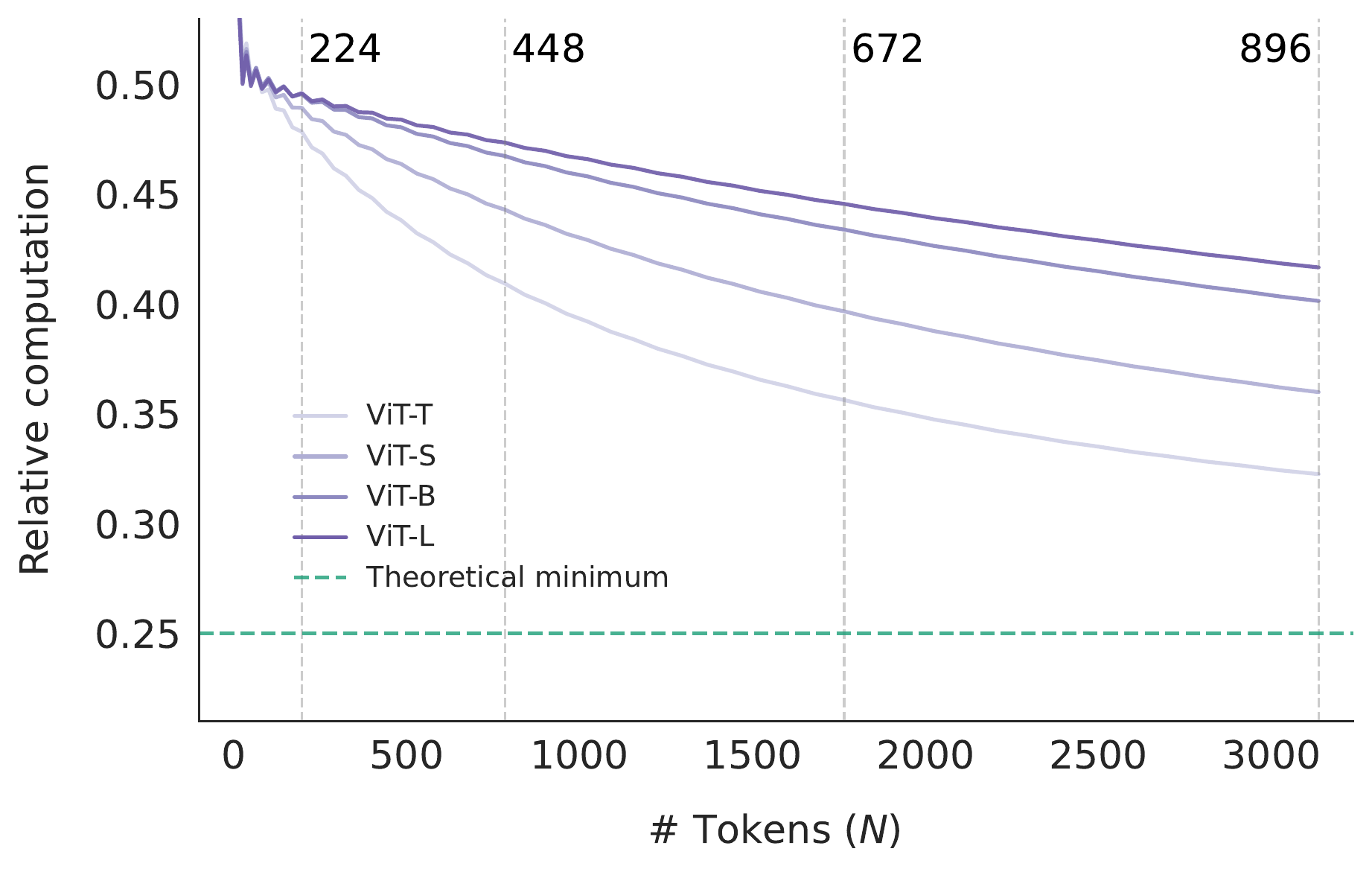}\vspace{3mm}\\
\textbf{Keep rate 0.25 (Theoretical)} &
\textbf{Keep rate 0.25 (Empirical)} &\vspace{2mm}
\\
\includegraphics[width=0.5\linewidth]{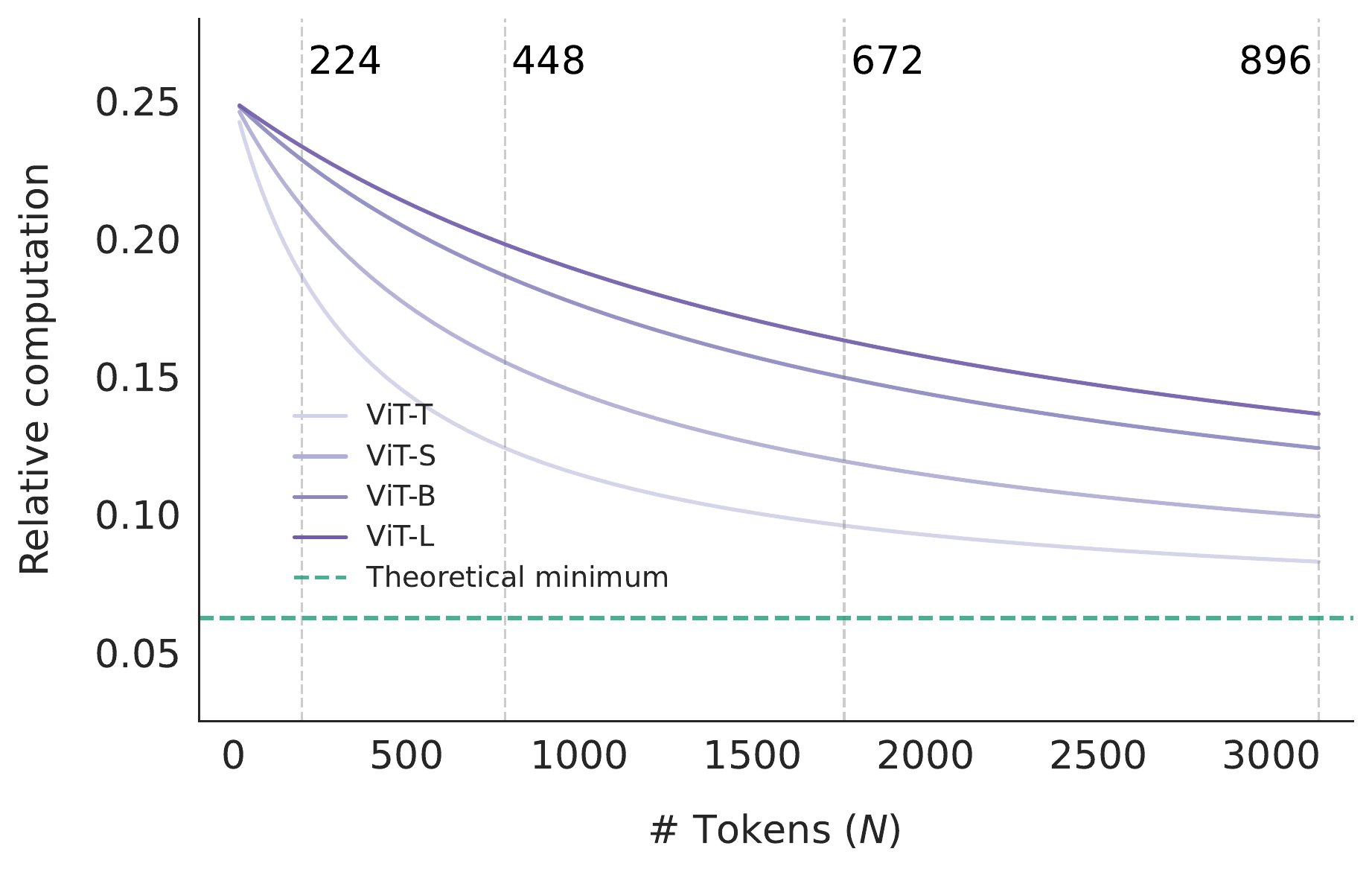} &
\includegraphics[width=0.5\linewidth]{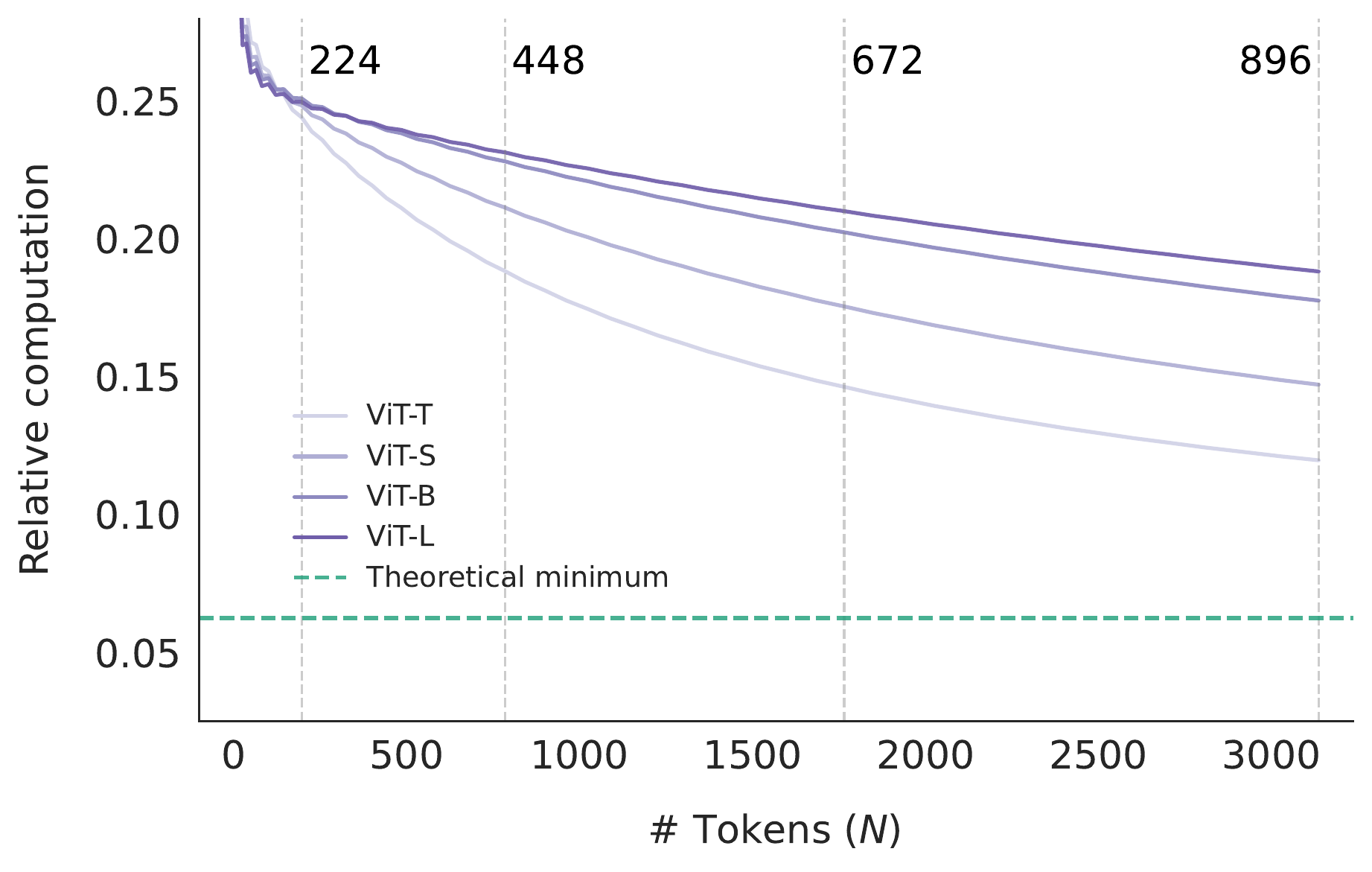}\\

\end{tabular}


\normalsize
\caption{
\emph{Computational savings with PatchDropout increase for larger sequence length $N$.} 
We illustrate the theoretical (left) and empirical (right) relative savings in computation when using PatchDropout with keep rates 0.5 and 0.25 for different input sequence lengths and ViT models.
Different sized images with a fixed patch size 16 result in different numbers of tokens (vertical dashed lines).
As discussed in Section \ref{section:complexity}, empirical savings do not always correspond to the theoretical analysis due to various factors.
However, the trend remains consistent: as the image size, and thus the number of input tokens $N$ increases, the observed computational savings approach the theoretical minimum $\lim_{N\to\infty}$.
} 
\label{fig:complexity}
\vspace{-4mm}
\end{figure}

In Figure~\ref{fig:complexity} we illustrate the relative drop in computations, both according to Eq. \ref{eq1} and empirically according to the number of FLOPs. 
We compare for different sequence lengths $N$ when using two keep rates of $0.5$ and $0.25$. 
As discussed above, the computational saving of PatchDropout increases with increasing number of tokens $N$. 
It can be seen that the drop in computation is similar to the keep rate for small $N$, but gradually converges to a quadratic savings with increasing $N$. 
While the theoretical and empirical trends are similar, the empirical saving converges slower due to the additional computations discussed above.   
Note that $N$ varies with image size $H\times W$ and patch size $P$. 
The default image size across multiple vision benchmark datasets is $224\times224$. 
At this scale, the relative computation stays close to the value of its keep rate -- the images are not large enough to benefit much from the quadratic savings. 
However, many real-world tasks demand high-resolution images, like the ones typically encountered in the medical domain. 
Here, we observe large computational savings as the sequence length increases. 

Finally, an important factor to note is that the embedding dimensions $d$ varies between different ViT variants, affecting their computational savings. 
In general, the relative computation of PatchDropout on smaller ViTs (with smaller $d$) decreases faster compared to larger ones.

\section{Experimental Setup}
\label{experiments}

We evaluate PatchDropout across a number of different ViT variants and datasets.
As representative ViT models, we selected \deits \cite{touvron2021training} of different capacity and \swins \cite{liu2021swin}, which are ViT variants that scale linearly with respect to the input sequence length by design.
For datasets, we selected three standard benchmark image classification datasets and a real-world medical dataset of high-resolution images.
Below, we describe the experimental settings in detail, and in Section \ref{results} we report our findings.


\paragraph{Data selection}
In this work we attempt to economize vision transformers such that they can be utilized for tasks where high-resolution images are necessary for accurate predictions.
To this end, we select a subset of 190,094 high-resolution images from \csaw, a population based cohort which consists of millions of mammography scans primarily developed for breast cancer tasks \cite{dembrower2019multi,matsoukas2020adding, sorkhei2021csaw}. 
Here, we focus on the breast cancer risk prediction, a sensitive classification task.
The data is split at the patient-level and the validation set contains balanced classes, resulting in 152,922 training images, 3,256 validation images, and 33,916 testing images. 
Furthermore, to validate the applicability of PatchDropout in other domains and on conventional image sizes, we run experiments on 3 standard image classification datasets: \imagenet \cite{deng2009imagenet}, \cifar \cite{krizhevsky2009learning} and \places \cite{zhou2017places}.
Adhering to standard practice, we report our results on the official validation splits of \imagenet and \places and we use 1\% of the training data for validation.
On \cifar, 2\% of the training images comprise the validation set and the results are reported on the official test set.



\paragraph{Preprocessing} 
Images from \csaw are in DICOM format, and require several pre-processing steps which are detailed below. 
Using the DICOM metadata, we re-scale the intensity values and correct any images with inverted contrast. 
Following  \cite{liu2020decoupling}, certain images are excluded according to a set of exclusion criteria.
The purpose is to filter out noisy images, images with implants, biopsies and mammograms with aborted exposure.
The mammograms with cancer signs are separated from those intended for risk estimation.
More specifically, cases with examination 60 days in advance of diagnosis are excluded in order to avoid risk conflation. 
For the other datasets we only resize their images to meet the needs of our experiments, no additional pre-processing was performed. 
Further details are provided in Supplementary \ref{preprocessing}.

\paragraph{Models and training protocols}
In this study, we primarily use \deits \cite{touvron2021training} which are similar in spirit and computational complexity to the original ViTs \cite{dosovitskiy2020image}. 
Unless otherwise specified, the model choice is \deitBase trained on $16\times16$ patches (denoted as \deitBase/16), and it is trained on input size $224\times224$.
Additionally, to show that PatchDropout is agnostic to architectural selections, we run ablations using \swins \cite{liu2021swin}.
\swins
designed to reduce the computational complexity of the original ViTs.
They scale linearly with respect to the input size and they inherit some of the CNN's inductive biases by design.
Additional implementation details can be found in Supplementary \ref{model_training}.

\section{Results and Discussion}
\label{results}

\begin{table}[t]
    \scriptsize
    \centering
        \begin{tabular}{@{}c@{\hspace{2mm}}c@{\hspace{2mm}}c@{\hspace{2mm}}c@{\hspace{2mm}} c@{\hspace{2mm}}@{}c}
        \toprule 
        &\textbf{Input} &\textbf{Keep rate} & \textbf{Memory (GB)} & \textbf{GFLOPs} & \textbf{AUC} \\
        \midrule
        &224 &1 & 1.46 & 17.58 & 64.71\% \\
        \midrule
        &896 &0.05 & 1.50  & 15.65 &  65.27\%\\
        &896 &0.10 & 1.65 & 30.37 &  65.59\%\\
        &896 &0.25 & 2.51 & 79.96 &  66.63\%\\
        &896 &0.50 & 5.15 & 180.64 &  67.03\%\\
        &896 &1 & 14.86 & 449.98 & 66.47\%\\
        \bottomrule
        \end{tabular}
    \caption{\emph{Performance (AUC), memory and compute savings using PatchDropout on \csaw.} The memory is computed with a batch size of 1 on a single GPU.}        
    \label{tab:csaw_keep_rates}
\vspace{-4mm}
\end{table}

We begin this section by demonstrating that not all input patches are necessary during training -- hence, we can randomly discard a large proportion of them.
Then, we show how PatchDropout can be used not only to save memory and compute but also to improve the model's predictive performance.
Finally, we analyze the regularization effects of PatchDropout and its role as an augmentation method.
Unless otherwise stated, each experiment is repeated 3 times and we report the mean value of the appropriate metric for each dataset.
For \imagenet, \cifar and \places, we report top-1 accuracy and for \csaw the exam-level AUC, where the predictions take the average score of each mammogram in an examination.


\begin{figure}[t!]
\centering
\scriptsize
\begin{tabular}{@{}c@{}c@{}}
\hspace{5mm}\imagenet &
\hspace{5mm}\cifar 

\\

\includegraphics[width=0.5\linewidth]{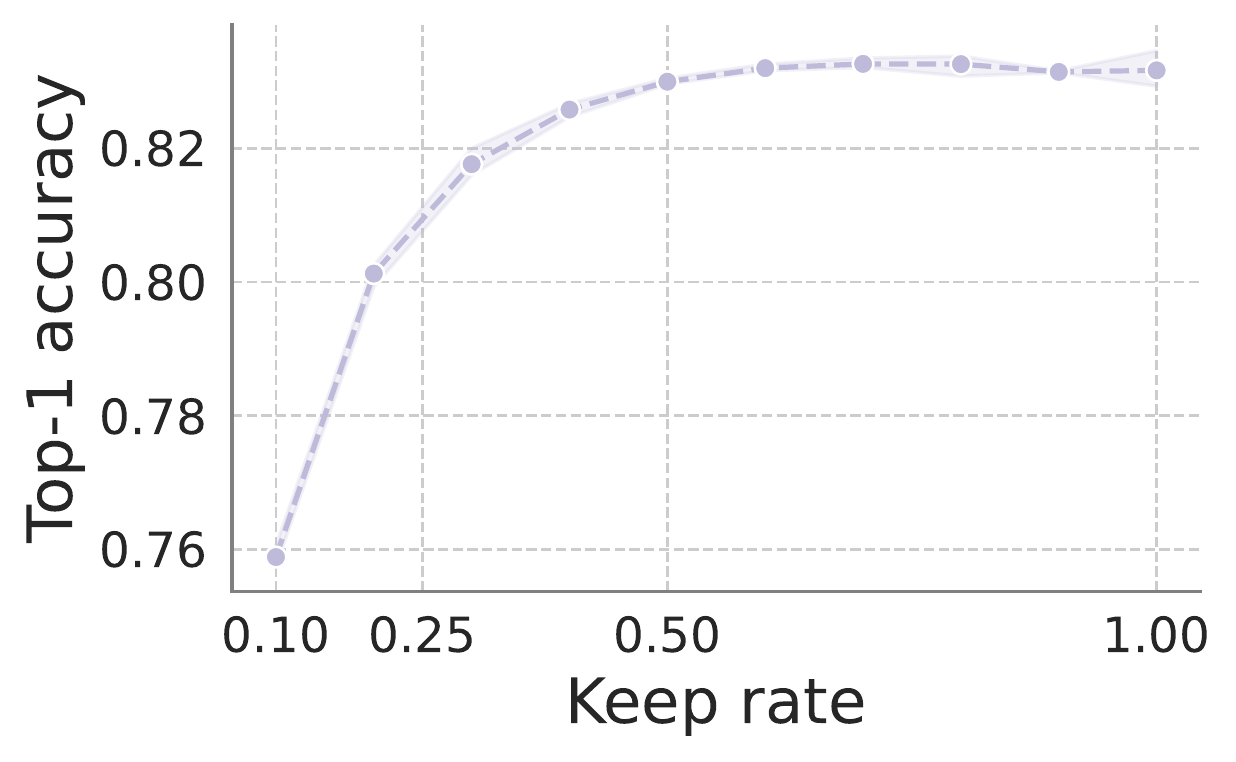} &
\includegraphics[width=0.5\linewidth]{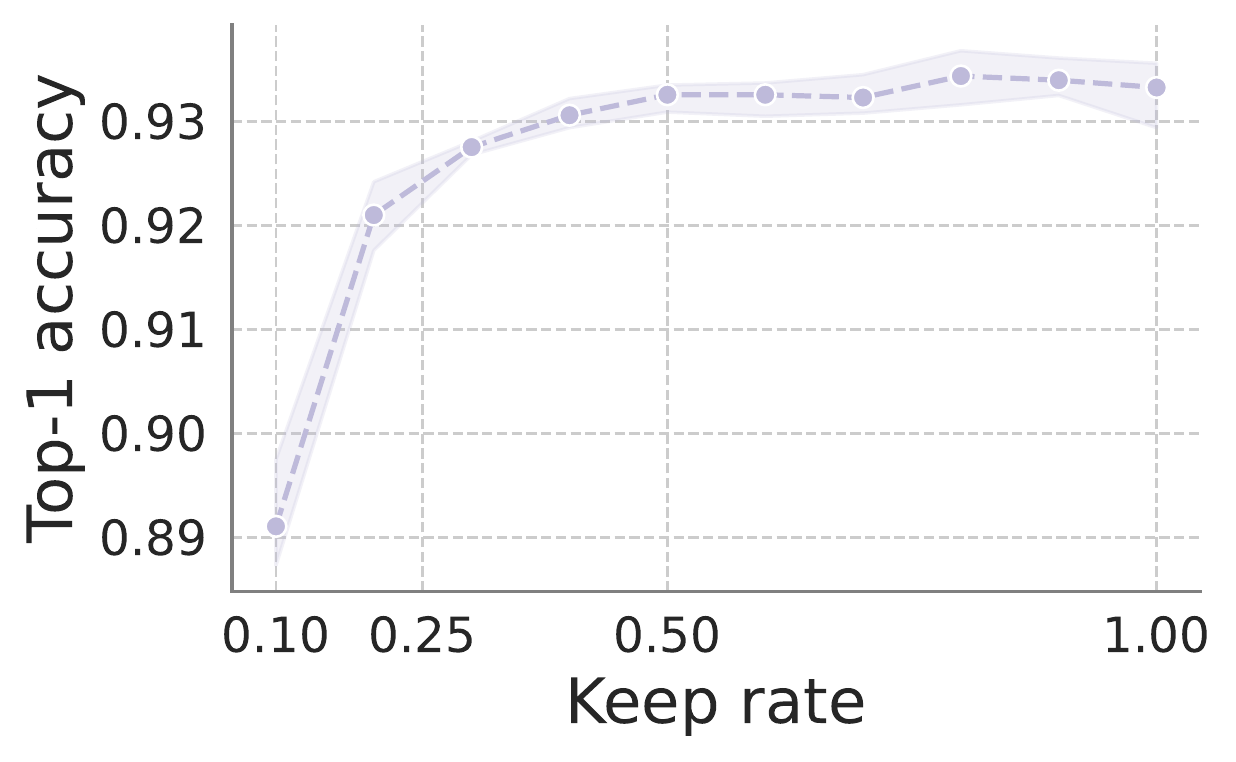} \\
\hspace{5mm}\places  &
\hspace{5mm}
\csaw
\\
\includegraphics[width=0.5\linewidth]{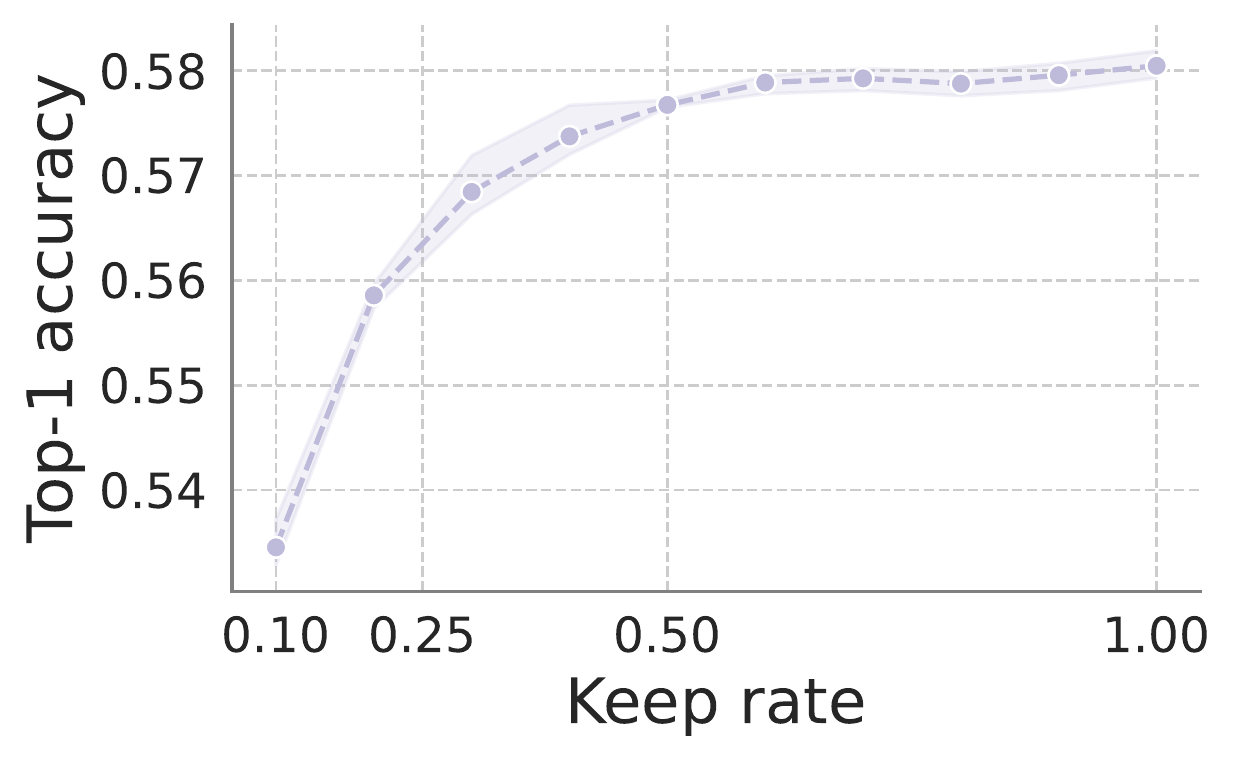} &
\includegraphics[width=0.5\linewidth]{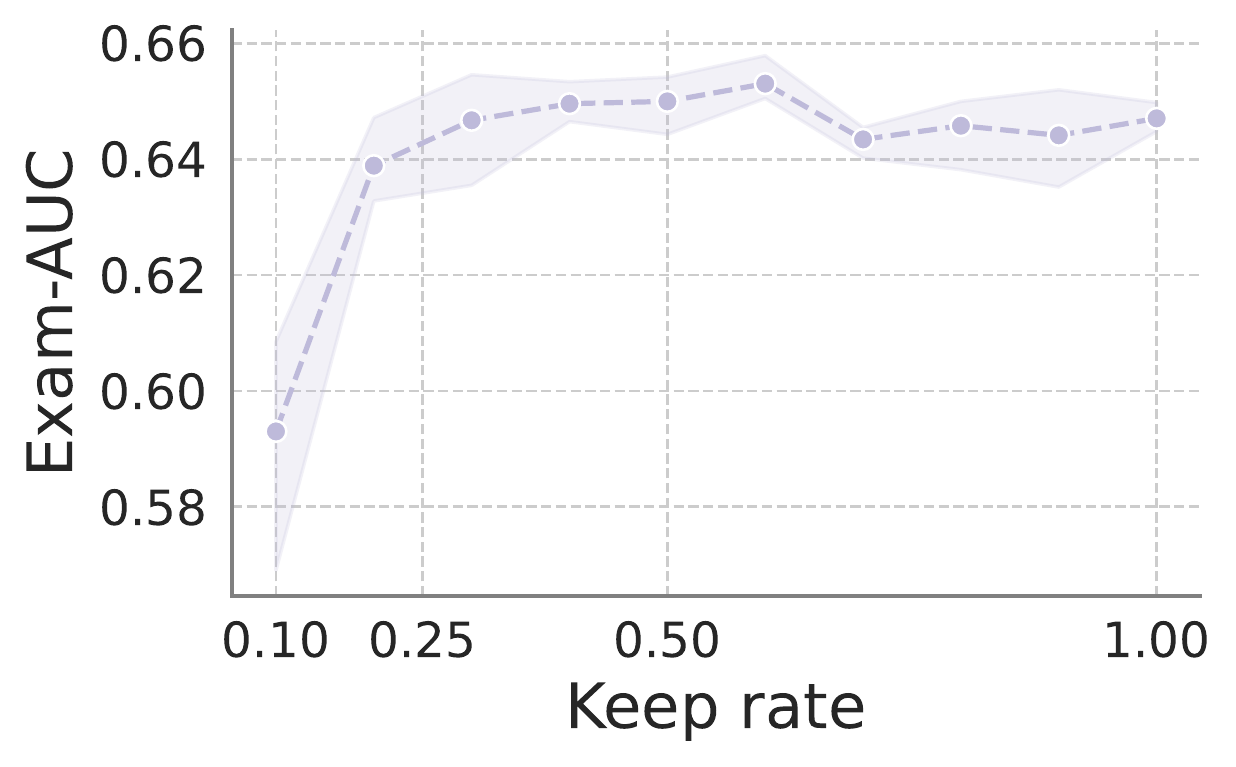} \\

\end{tabular} 
\vspace{-2mm}

\caption{\emph{Not all input patches are necessary to be present.} 50\% of the input patches are sufficient to preserve model performance for image size $224\times224$: it improves efficiency $2\times$ while the performance drop is contained at only 0.17\% on \imagenet, 0.07\% on \cifar and 0.38\% on \places. On \csaw, keeping around half of the input patches results in 0.25\% - 0.60\% increase in AUC compared with keeping all tokens. }
\label{fig:main_results}
\end{figure}

\addtolength{\tabcolsep}{-2.5mm}  
\begin{table}[t]
\footnotesize

\centering
\scalebox{0.9}{
\begin{tabular}{@{}c@{\hspace{1mm}}c@{\hspace{1mm}}c@{\hspace{1mm}} c@{\hspace{1mm}}c@{\hspace{1mm}}c@{\hspace{1mm}}c@{\hspace{1mm}}c@{\hspace{1mm}}c}
        \toprule
        & \makecell{\textbf{Keep} \\ \textbf{rate} }& \makecell{\textbf{Memory} \\ \textbf{(GB)}}  & \textbf{GFLOPs} & \textbf{\imagenet} & \textbf{\cifar} & \textbf{\places}& \textbf{\csaw} \\
        \midrule
        & 1   & 20.96 & 17.58 & 83.17\%  & 93.33\% & 58.05\% & 64.71\% \\
        \midrule
        & 0.9 & 0.89$\times$ & 0.90$\times$ & $-$0.03\%  & $+$0.07\% & $-$0.09\%& $-$0.30\% \\
        & 0.8 & 0.78$\times$ & 0.79$\times$ & $+$0.09\%  & $+$0.11\% & $-$0.17\%& $-$0.13\% \\
        & 0.7 & 0.68$\times$ & 0.69$\times$ & $+$0.09\%  & $-$0.10\% & $-$0.12\% & $-$0.37\% \\
        & 0.6 & 0.57$\times$ & 0.59$\times$ & $+$0.03\%  & $-$0.07\% & $-$0.16\% & $+$0.60\% \\
        & 0.5 & 0.48$\times$ & 0.50$\times$ & $-$0.17\%  & $-$0.07\% & $-$0.38\% & $+$0.29\%  \\
        & 0.4 & 0.39$\times$ & 0.40$\times$& $-$0.59\%  & $-$0.27\% & $-$0.68\% & $+$0.25\% \\
        & 0.3 & 0.30$\times$& 0.40$\times$ & $-$1.41\%  & $-$0.58\% & $-$1.21\% & $-$0.04\% \\
        & 0.2 & 0.22$\times$ & 0.20$\times$ & $-$3.04\%  & $-$1.23\% &$-$2.19\% & $-$0.82\% \\
        & 0.1 & 0.14$\times$ & 0.10$\times$ & $-$7.28\%  & $-$4.22\% & $-$4.60\%& $-$5.41\% \\
        \bottomrule
\end{tabular} }
\caption{
\emph{Performance, memory and compute savings using PatchDropout on various datasets with $224\times224$ images.} 
}
\label{tab:main_tab}
\vspace{-3mm}
\end{table}

        


\paragraph{Are all input patches necessary during training?}
To assess the impact of PatchDropout and determine whether all tokens are necessary for training ViTs, we conduct experiments where different percentages of the tokens are presented to the model.
As illustrated in Figure~\ref{fig:fig1} and Table~\ref{tab:csaw_keep_rates}, 25\% of the tokens are enough to train an accurate model on high-resolution \csaw images of $896\times896$ pixels, while consuming more than 80\% less memory and compute.
Interestingly, models trained with 25\% or 50\% of the tokens outperform a model that uses all tokens. This hints at a regularization effect for PatchDropOut that we will discuss later.

In Figure~\ref{fig:main_results} and Table~\ref{tab:main_tab} we explore how this trend translates to standard image classification benchmark datasets using $224\times224$ images.
We observe that performance varies as a function of the keep rate.  
In all cases, a keep rate of 50\% or larger is sufficient to maintain good performance. 
When using exactly 50\% of the tokens, the performance drop is contained to only 0.17\% on \imagenet, 0.07\% on \cifar and 0.38\% on \places. 
The reduction in memory and computation are significant and similar to the keep rate.

\begin{table}[t] 

    \scriptsize
    \centering
\begin{tabular}{@{}c@{\hspace{1.5mm}}c@{\hspace{1.5mm}} c@{\hspace{1.5mm}}c@{\hspace{1.5mm}}c@{\hspace{1.5mm}}c@{\hspace{1.5mm}}c@{\hspace{1.5mm}}c@{\hspace{1.5mm}}@{}c}
        \toprule
        & \textbf{Input}& \textbf{Patch} & \textbf{Keep rate}&\textbf{GFLOPs} & \textbf{\imagenet} & \textbf{\cifar} & \textbf{\csaw} \\
        \midrule
        &64 & 16 & 1 & 1.46  & 66.78\%  &87.27\% &-\\
        &64 & 8 & 0.25 & 1.46  & 70.57\% &89.77\% &-\\
        &128 & 16 & 0.25 & 1.49  & \textbf{76.25\%} &\textbf{91.30\%} &-\\
        \midrule
        &112 & 16 & 1  & 4.33 &77.65\% & 91.98\% & 63.07\%\\
        &112 & 8 & 0.25 & 4.33 & 79.11\% & 92.38\% & 60.08\%\\
        &224 & 16 & 0.25 & 4.41 & \textbf{81.02\%} & \textbf{92.50\%} &\textbf{64.87\%}\\
        \midrule
        &224 & 16 & 1 &   17.58 & 83.17\% & \textbf{93.33\%} & 64.71\%\\
        &224 & 8 & 0.25 & 17.58 & \textbf{83.43\%} & 92.71\% & 64.28\%\\
        &448 & 16 & 0.25 & 17.93 & 83.26\% & 92.20\% & \textbf{65.59\%}\\
        \midrule
        &448 & 16 & 1 &  78.57 &-& -& 66.31\%\\
        &448 & 8 & 0.25 &  78.57 &-&- & 66.13\%\\
        &896 & 16 & 0.25& 79.96 &-&- & \textbf{66.63\%}\\
        \bottomrule
        \end{tabular}
\caption{\emph{Effect of varying image size and patch size.} The impact in terms of FLOPS and performance  of changing the image size and patch size is measured using PatchDropout over multiple datasets.}        
\label{tab:token_number}
\vspace{-2.5mm}
\end{table} 

\paragraph{Can we trade the savings in memory and compute introduced by PatchDropout
for more accurate predictions?
}
In the previous analysis we saw that PatchDropout allows for significant memory and computational savings without compromising the model's performance. This saving can enable a more elaborate model selection (\eg finer grid search) or a wider range of training choices (\eg larger batch size) or a more accurate but computationally-heavy architecture.
Therefore, the next question we ask is whether we could utilize the saved memory and compute to improve the model's predictive performance while keeping the computational budget similar to the one used for the full token sequence.
Our experiments show that this can be easily achieved with two simple design choices:
\begin{enumerate}[label={(\arabic*)}]
    \vspace{-1mm}
    \item Increasing the total token sequence by (a) using higher resolution images, or (b) decreasing the patch size.
    \vspace{-1.5mm}    
    \item Employing models with greater capacity.
    \vspace{-1mm}    
\end{enumerate}





\vspace{-3mm}
\subparagraph{ -- Larger images or smaller patch size}
It has been proven that ViTs perform better on larger images and smaller patch sizes \cite{dosovitskiy2020image, touvron2021training, caron2021emerging}.
However, this comes with a large memory and computational overhead due to the increased input sequence length, as described in Section \ref{methods}.
PatchDropout mitigates this cost by reducing the sequence length, allowing for the utilization of larger images and smaller patches.
Table \ref{tab:token_number} illustrates the trade-off between the model's performance and the input sequence length for different settings on various datasets.

Trading the saved compute from PatchDropout for larger images yields a large performance boost for almost all setups (compare the \textit{1$^{st}$} and \textit{3$^{rd}$} row of each group).
For example, comparing \imagenet for images of size $128\times128$ with keep rate of 0.25 with the $64\times64$ images with all tokens retained, we find that this simple trade-off results in a nearly $10\%$ absolute increase in accuracy for similar cost.
The performance gains lessen, however, for larger computational budgets.
This trend is observed across all the datasets.

Trading the savings obtained using PatchDropout for smaller patch sizes also yields significant performance gains.
However, the gains are not as consistent as for resolution (compare the \textit{1$^{st}$} and \textit{2$^{nd}$} row of each group).
For the natural domain, smaller patch sizes improve model performance, as expected from  \cite{dosovitskiy2020image, touvron2021training}.
On \csaw, smaller patch size seems to negatively affect performance, but the effect diminishes as we move to higher resolutions.

Note that, in some cases, images are up-sampled in our experiments.
In general, we notice that higher resolution and smaller patch size is usually beneficial, but not always. 
We speculate that as we move away from a dataset's native resolution, larger input size and smaller patch sizes might have a negative impact on model performance due to significant information loss.
Nevertheless, PatchDropout allows for the exploration of hyperparameter settings unfeasible to reach with the full token sequence.

\begin{table}[t]
    \scriptsize
    \centering
    
        \begin{tabular}{@{}c@{\hspace{1mm}}c@{\hspace{2mm}} c@{\hspace{2mm}}c@{\hspace{1mm}}c@{\hspace{1mm}}c@{\hspace{1mm}}c@{\hspace{1mm}}c@{\hspace{1mm}}c@{\hspace{1mm}}@{}c}
        \toprule
        & \textbf{Model}  &\makecell{\textbf{Keep} \\ \textbf{rate}}& \makecell{\textbf{Memory} \\ \textbf{(GB)}}& \textbf{GFLOPS} & \textbf{\imagenet} & \textbf{\cifar} & \textbf{\csaw}\\
        \midrule
        & \deitTiny & 1 & 5.06 & 1.26  & 75.22\%  &86.94\% & 63.45\%\\
        & \deitsmall & 0.25 & 2.46 & 1.15  & \textbf{78.09\%} & \textbf{90.30\%} & \textbf{63.76\%}\\
        \midrule
        & \deitsmall & 1 & 10.20 & 4.61  & 80.69\% & 91.08\% & 64.62\%\\
        & \deitBase & 0.25 & 5.46 & 4.41 & \textbf{81.02\%}& \textbf{92.50\%} & \textbf{64.87\%}\\
        \midrule
        & \deitBase & 1 & 20.96 & 17.58  & 83.17\% & 93.33\% & 64.71\%\\
        & \deitLarge & 0.25 &15.34 & 15.39 & \textbf{83.81\%}& \textbf{93.97\%} & \textbf{65.31\%}\\
        
        \bottomrule
        \end{tabular}
\caption{\emph{Impact of training larger ViT variants with PatchDropout using $224\times224$ images.}}
\label{tab:model_variant}
\end{table} 

\begin{table}[t]
    \scriptsize
    \centering
        \begin{tabular}{@{}c@{\hspace{1mm}}c@{\hspace{1mm}} c@{\hspace{1mm}}c@{\hspace{1mm}}c@{\hspace{1mm}}c@{\hspace{1mm}}c@{\hspace{1mm}}c@{\hspace{1mm}}c@{\hspace{1mm}}@{}c}
        \toprule
        & \textbf{Model} & \textbf{Depth}  &\makecell{\textbf{Keep} \\ \textbf{rate}}& \makecell{\textbf{Memory} \\ \textbf{(GB)}} &  \textbf{GFLOPS} & \textbf{\imagenet} & \textbf{\cifar} & \textbf{\csaw}\\
        \midrule
        & \deitBase & 12  & 1 &20.96 & 17.58  & \textbf{83.17\%} & 93.33\% & 64.71\%\\
        & \deitBase & 24  & 0.5 &19.73& 17.31 & 83.06\% & \textbf{93.40\%} & \textbf{65.42\%}\\
        & \deitBase & 48  & 0.25 &20.95& 17.31 & 81.46\% & 92.71\% & 65.31\% \\
        \bottomrule
        \end{tabular}
\caption{\emph{Impact of training deeper models with PatchDropout using $224\times224$ images.}}
\label{tab:model_depth}
\vspace{-3mm}
\end{table} 

\vspace{-2.5mm}
\subparagraph{-- Models with larger capacity}
Increasing the model capacity is another way to attain better predictions. 
The memory and compute saved from PatchDropout can be spent on training larger ViT variants.
In Table \ref{tab:model_variant}, we explore this trade-off. Interestingly, the trend is that larger models using PatchDropout are consistently better than the smaller variants of equivalent cost that use all tokens. 
This trend follows across all data domains, with memory efficiency improving up to $2.1 \times$.
Natural datasets win bigger gains with PatchDropout using larger models, as compared to increased image size or reducing token size.

An alternative way to increase the model capacity is to stack more transformer blocks to a particular ViT variant.
We explore this trade-off for a fixed computational budget by increasing the model's depth and varying the keep rate.
We report the results in table~\ref{tab:model_depth}.
When doubling the model's transformer blocks with PatchDropout we observe performance boosts for \cifar and \csaw.
However, we note worse performance on \imagenet when the model becomes too deep. We attribute this to the fact that ViT architectures are optimized for \imagenet.

Ensembles of models are yet another alternative to obtain more accurate predictions.
We explore how computational savings from PatchDropout can be spent training additional models for use in an ensemble.
In Table \ref{tab:ensemble}, we show that an ensemble of two networks trained with 50\% keep rate consistently outperforms a single model without PatchDropout.
However, the gains diminish when lower keep rates (25\%) are traded for an ensemble for four networks.

\begin{table}[t!]
    \scriptsize
    \centering
    \begin{tabular}{@{}c@{\hspace{1.5mm}} c@{\hspace{1.5mm}}c@{\hspace{1.5mm}}c@{\hspace{1.5mm}}c@{\hspace{1.5mm}}c@{\hspace{1.5mm}}c@{\hspace{1.5mm}}c}
        \toprule
        & \textbf{\#Networks} & \textbf{Keep rate}  & \textbf{GFLOPs} & \textbf{\imagenet} & \textbf{\cifar} & \textbf{\csaw} \\
        \midrule
        & 1 & 1 & 17.58 & 83.17\% & 93.33\% & 64.71\%\\
        & 2 & 0.5 & 17.44 & \textbf{83.48\%} & \textbf{93.74\%} & \textbf{65.26\%}\\
        & 4 & 0.25 & 17.66 & 82.20\% & 93.45\% & 64.92\%\\
        \bottomrule
        \end{tabular}
    \caption{\emph{Using PatchDropout compute savings to train an ensemble.}}
    \label{tab:ensemble}
    \vspace{-4mm}
\end{table}

\paragraph{Can PatchDropout be used as a regularisation method?}
Previously, we noticed that PatchDropout can result in increased performance compared to the same settings but with all tokens, \eg in Figure~\ref{fig:main_results}. 
This has some implications about its regularization effects.
Thus we ask: \textit{(i)} Can PatchDropout be used as a regularizer? and \textit{(ii)} Does PatchDropout provide robustness against information removal?

To answer the first question we run experiments treating PatchDropout as an augmentation method.
In detail, at each iteration we uniformly sample a keep rate between 0.5 and 1 which we use to randomly select a subset of image patches.
We report the results in Table~\ref{tab:regularisation} and we conclude that PatchDropout is a useful augmentation method.
It provides regularization in the sense that generalization is improved across all the datasets.
This is not entirely surprising, as PatchDropout behaves similarly to known CNN regularization methods, like cutout \cite{devries2017improved}.

\begin{figure}[t]
\centering
\scriptsize

\begin{tabular}{@{}c@{}c@{}}
\hspace{5mm}\imagenet &
\hspace{5mm}\csaw
\\
\includegraphics[width=0.5\linewidth]{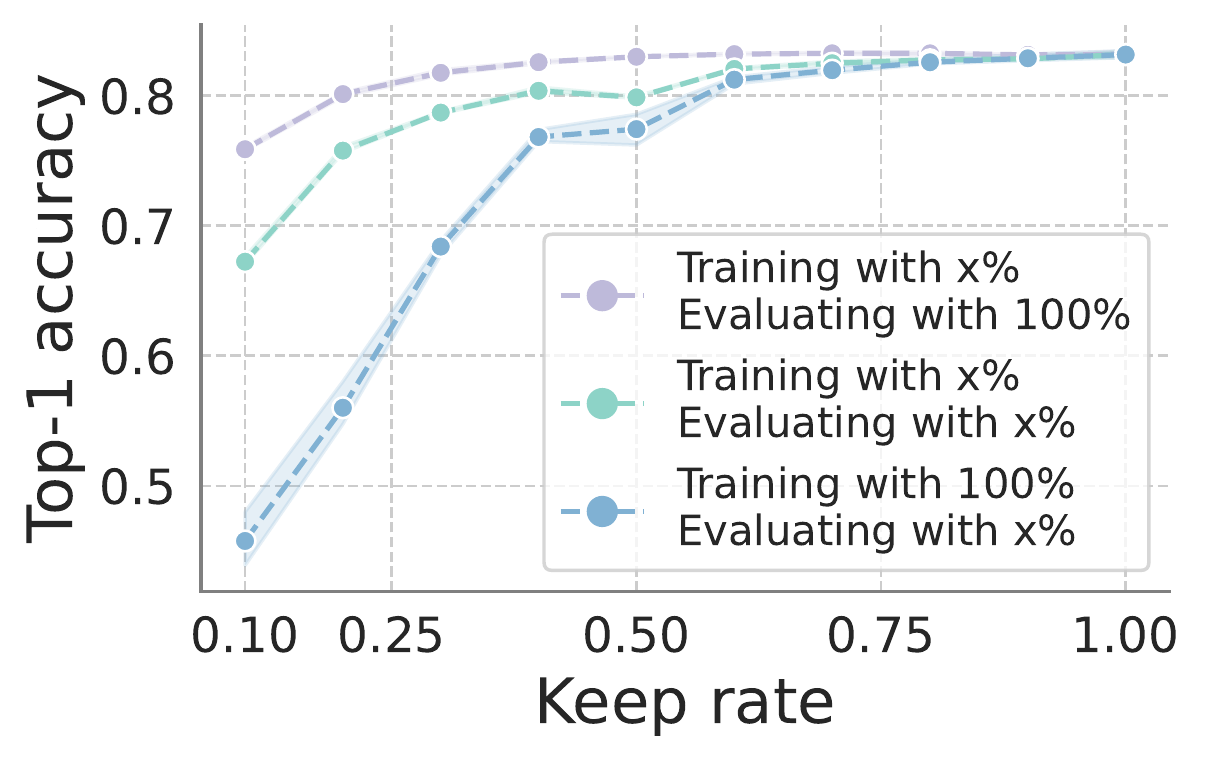}
 &
\includegraphics[width=0.5\linewidth]{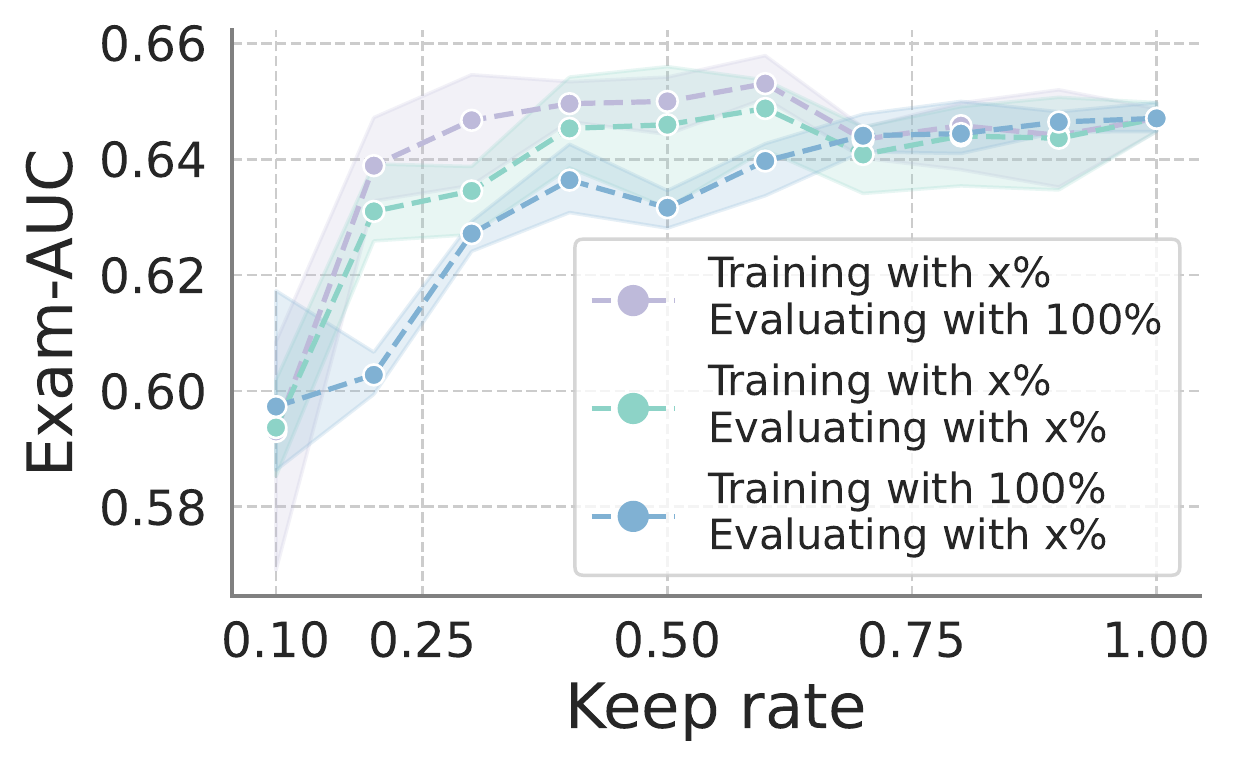}
\end{tabular}
\normalsize
\caption{
\emph{PatchDropout improves model robustness.}
We deny information to the model during inference by randomly dropping input patches and measuring the change in performance.
The green curve shows the model's performance when training with a percentage of the input patches and evaluating using the same keep rate.
The blue curve represents the model's performance when training using all patches but evaluating on a subset of the input patches.
For completeness, the purple curve shows training with PatchDropout and inference with all patches.
When 50\% or more of the patches are kept, this results in a minimal performance drop on \imagenet and increased predictive performance on \csaw.
The trends show that models trained with PatchDropout are more robust to missing information during inference.
}

\label{fig:robustness}
\vspace{-4mm}
\end{figure}

If PatchDropout has some regularization benefits, a further question is: \textit{can it provide robustness against information loss?}
To address this question, we evaluate models that have been trained with all image patches and models that have been trained using PatchDropout with different keep rates.
During test time, we randomly remove image content using different keep rates.
Our results are presented in Figure~\ref{fig:robustness}.
We find that, in all cases, models that have been trained with PatchDropout exhibit increased robustness against information removal (the green curve is consistently above the blue curve).
For completeness, we also report the curves when using all tokens at test time for models that have been trained with PatchDropout (purple curve in Figure~\ref{fig:robustness}).
These results further validates the regularization effects of our method.

\begin{table}[t]
    \scriptsize
    \centering
    \begin{tabular}{@{}c@{\hspace{3mm}} c@{\hspace{3mm}}c@{\hspace{3mm}}c@{\hspace{3mm}}c@{\hspace{3mm}}c}
        \toprule
        & \textbf{Keep rate} & \textbf{\imagenet} & \textbf{\cifar} & \textbf{\csaw} \\
        \midrule
        & 1 & 83.17\% & 93.33\% & 64.71\%\\
        & \{0.5,1\} & \textbf{83.32\%} & \textbf{93.57\%} & \textbf{65.04\%}\\
        \bottomrule
        \end{tabular}
    \caption{\emph{PatchDropout has regularization properties}. Rather than using all tokens, training ViTs with PatchDropout using random keep rates improves generalization. 
    }
    \label{tab:regularisation}
\end{table}  

\paragraph{Is PatchDropout constrained by the architectural choice?} 
Throughout our work we used the \deit model family as they are the most suitable for the purpose of our analysis.
This however, raises the question of whether PatchDropout is effective for other architectural choices.
To answer this question we run experiments using \swins \cite{liu2021swin} which are models purposely designed to reduce the computational complexity of \deits.
\swins operate using a window shifting approach and re-assignment of positional embeddings at each block.
This, necessitates a slightly different implementation of PatchDropout.
Instead of randomly sampling image patches, we apply structured sampling where we randomly sample column and row indices for each window to obtain the intersection tokens. 
This maintains the spatial relationships between tokens and enables smooth window shifting for \swins. 
The corresponding relative positional biases are sampled accordingly. 

\begin{figure}[t]
\centering
\scriptsize
\begin{tabular}{@{}c@{}c@{}}
\hspace{5mm}\imagenet &
\hspace{5mm}\csaw 
\\
\includegraphics[width=0.5\linewidth]{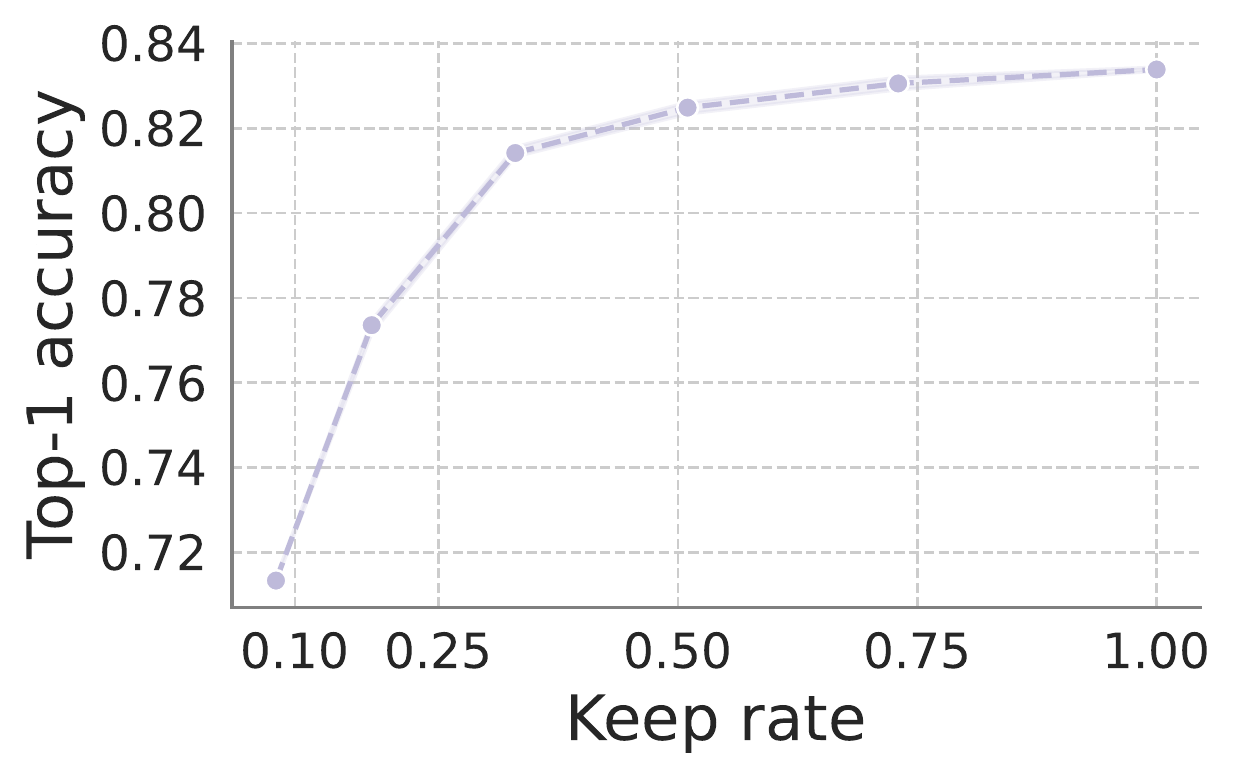}
 &

\includegraphics[width=0.5\linewidth]{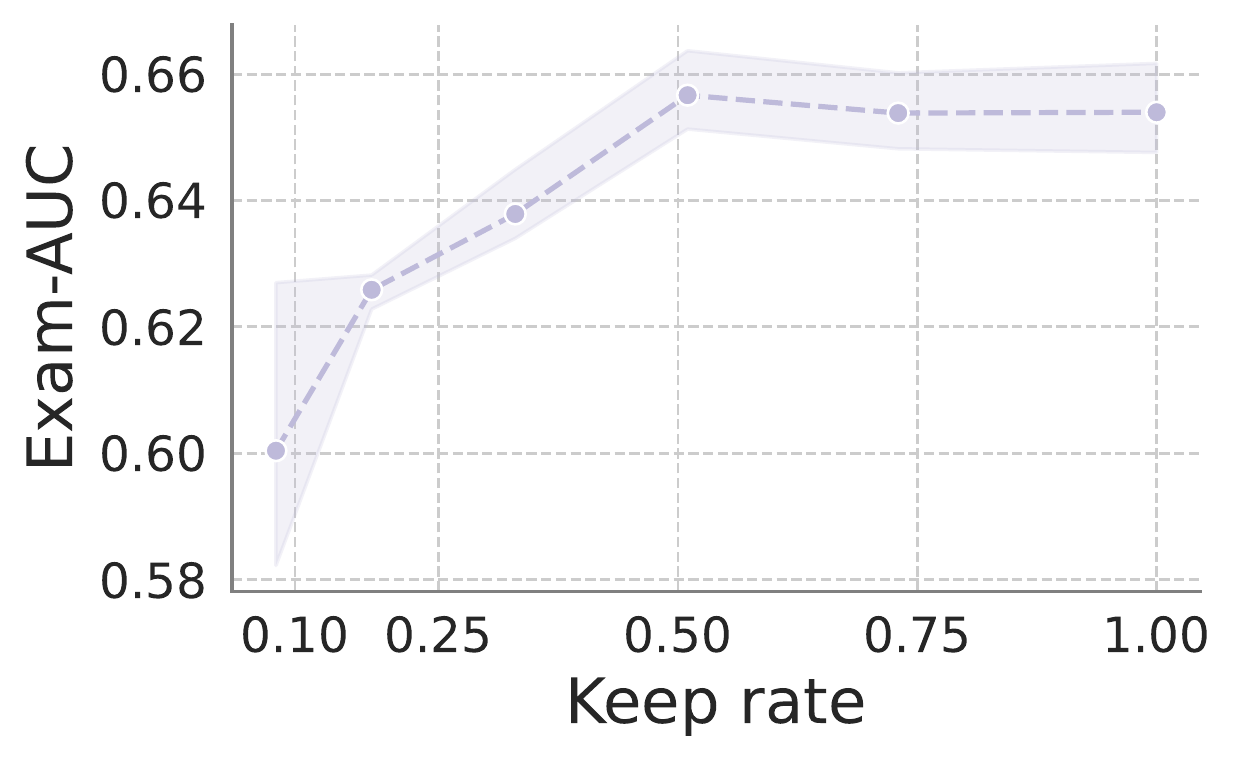}
\end{tabular}
\normalsize
\caption{\emph{PatchDropout also works for SWINs.} Despite their linear scaling via the reintroduction of CNN inductive biases, PatchDropout can be applied to \swins with a keep rate of 0.5 or higher without lowering performance.}
\label{fig:swin}
\end{figure}

We report our findings in Figure~\ref{fig:swin} when using $224\times224$ images for both \imagenet and \csaw.
We discern similar patterns with the ones in Figure~\ref{fig:main_results}.
\csaw exhibits small performance gains when using PatchDropout with keep rates larger than 50\% while \imagenet displays a 1\%  drop at 50\% keep rate.
The applicability of PatchDropout however is still valid, even for this architecture which has been developed to economize \deits by design.




\begin{figure}[t]
\centering
\scriptsize
\begin{tabular}{@{}c@{\hspace{1mm}}c@{\hspace{1mm}}c@{\hspace{1mm}}c@{\hspace{1mm}}c@{\hspace{1mm}}c@{}}
   & Original &
   Uniform & 
   Structured& 
   Cropping& 
   Random 
\\
&\includegraphics[width=0.17\linewidth]{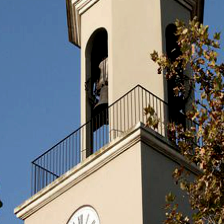} &
\includegraphics[width=0.17\linewidth]{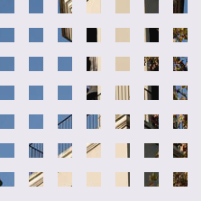} &
\includegraphics[width=0.17\linewidth]{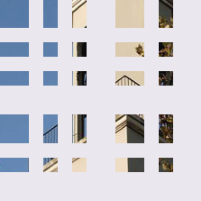} &
\includegraphics[width=0.17\linewidth]{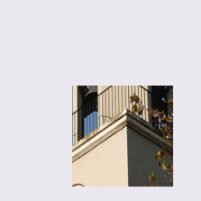} &
\includegraphics[width=0.17\linewidth]{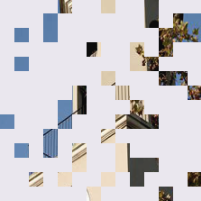} \\
  & 93.33\% &
  91.84\% &
  92.94\% &
  92.46\% &
  92.50\%

\end{tabular}
\normalsize
\caption{\emph{Impact of different patch sampling strategies}. Accuracy on \cifar is reported for keep rate 0.25 at  $224 \times 224$  resolution for various patch sampling strategies.}
\label{fig:sampling}
\vspace{-3mm}
\end{figure}


\vspace{-1mm}
\paragraph{Other ablation studies}
We conclude our analysis with two ablation studies aiming to assess the effectiveness of random sampling and the role of large scale pretraining.
\vspace{-2mm}
\subparagraph{-- How important is the sampling strategy in PatchDropout?}
To assess the efficacy of random sampling, which we use in our method, compared to other sampling methods, we conduct a small ablation study on \cifar where we train models using a fixed keep rate of 0.25 but we change the sampling method.
In Figure \ref{fig:sampling} we illustrate the sampling methods we used and we report the top-1 accuracy.
The results show that the effectiveness of the proposed method is not heavily dependent on the choice of sampling strategy. 
This, along with the results from \swins where we used structured sampling (see previous paragraph) indicates that PatchDropout is a general approach which can be easily incorporated into other types of ViT models.


\vspace{-2mm}
\subparagraph{-- Is PatchDropout sensitive to the initialization strategy?}
Throughout this work we utilized models pretrained on \imagenetk as vision transformers rely on large-scale pretraining, especially when working with small datasets \cite{dosovitskiy2020image, touvron2021training, matsoukas2021time, matsoukas2022makes}.
But, \textit{is PatchDropout useful when random initialization \cite{he2015delving} is used?}
To answer this question we train randomly initialized models on \csaw and we report the results in Table~\ref{tab:rand_init}.
Indeed, PatchDropout works with randomly initialized models on \csaw, although with diminishing performance gains, suggesting that the proposed method is agnostic to the initialization strategy.  



\begin{table}[t]
    \scriptsize
    \centering
    \begin{tabular}{@{}c@{\hspace{3mm}} c@{\hspace{3mm}}c@{\hspace{3mm}}@{}c}
        \toprule
        & \textbf{Keep rate}  & \textbf{\imagenetk init.} &\textbf{Random init. } \\
        \midrule
        & 1 & 64.71\% & 59.32\% \\
        & 0.50 & $+$0.29\% & $+$0.16\% \\
        & 0.25 & $+$0.16\% & $-$0.90\%\\
        \bottomrule
        \end{tabular}
    \caption{\emph{Impact of \imagenetk initialization for PatchDropout on \csaw.}}
    \label{tab:rand_init}
    \vspace{-4mm}
\end{table}

\section{Conclusion}
In this work, we rely on the fact that the spatial redundancy encountered in image data can be leveraged to economize vision transformers and we propose a simple yet efficient method, PatchDropout.
By dropping input tokens at random, our method results in significant memory and computation reduction, especially on high-resolution images.
In addition, we demonstrate how the saved compute introduced by PatchDropout can be exchanged for better predictive performance under the same memory and computational budget.
Finally, we show that PatchDropout can act as a regularization technique during training, resulting in increased model robustness.
PatchDropout requires minimal implementation and works with off-the-shelf vision transformers.
We believe that PatchDropout should be an essential tool in every practitioner’s toolkit 
to reduce the memory and computational demands in transformer training.


\paragraph{Broader Impact}

Reducing computational requirements can help to democratize deep learning by making it cheaper to train models. 
Achieving an equitable outcome for vulnerable and disadvantaged groups, who might lack access to sufficient funding and resources -- including but not limited to small academic groups, hospitals, and companies, necessitates a multitude of solutions. 
Apart from social benefits, one might consider the positive impacts with regards to climate as well. 
Data centers worldwide account for a substantial portion of energy consumption and GHG emissions which can be mitigated by shrinking computation during model development if applied widely. Despite our efforts, training state-of-the-art networks such as ViTs is still computationally expensive and thus has significant carbon footprints.


\paragraph{Acknowledgements.}
This work was partially supported by MedTechLabs (MTL), the Swedish Research Council (VR) 2017-04609, Region Stockholm HMT 20200958, and Wallenberg Autonomous Systems Program (WASP). The computations were enabled by the Berzelius resource provided by the Knut and Alice Wallenberg Foundation at the National Supercomputer Centre.

{\small
\bibliographystyle{ieee_fullname}
\bibliography{reference}
}

\clearpage
\twocolumn[
{\center\baselineskip 18pt
    \vskip .25in{\Large\bf
    Supplementary Material for PatchDropout: Economizing Vision Transformers Using Patch Dropout \par
}\vskip .52in}
]
\begin{appendices}
\section{Preprocessing} \label{preprocessing}
The images from CSAW are resized to $896\times896$, $448\times448$, $224\times224$ and $112\times112$ for the purposes of this work.
The image resolution on \imagenet varies and has an average resolution of $469\times387$.
For \places we used images of size $256\times256$ and for \cifar $32\times32$.
We resize these to $448\times448$, $224\times224$, $128\times128$, $112\times112$ and $64\times64$ in our experiments using bi-linear interpolation. 

\section{Models and training protocols} \label{model_training}



All models are initialized from \imagenetk  pre-trained weights and subsequently fine-tuned on the target task. 
Hyper-parameters are selected through grid search based on the result from the validation set.
We use early stopping. 
The batch size is consistently 128 for all experiments, and each model is trained with an SGD optimizer with momentum set to 0.9.
In every experiment, a linear learning rate warmup is utilized for the first 2 epochs. 
The learning rates are $3\times10^{-4}$ on \imagenet, $5\times10^{-4}$ on \cifar and \places, and $10^{-4}$ on \csaw, following the results of our grid search.
We use a weight decay of $1\times10^{-4}$ and label smoothing \cite{szegedy2016rethinking} of 0.1. 
We apply randomly resized cropping, random horizontal flipping, random rotation and color jittering as augmentations.
Unless otherwise specified, each experiment was repeated three times and we report the mean.
All the experiments are conducted with PyTorch \cite{NEURIPS2019_9015}. The number of FLOPs is counted using the fvcore toolkit \cite{fvcore}, and the allocated memory is calculated on a single GPU with a batch size of 128, unless otherwise stated.

\end{appendices}

\end{document}